\documentclass[letterpaper, 10 pt, conference]{ieeeconf}  
\usepackage{graphicx}
\usepackage[caption=false,font=normalsize,labelfont=sf,textfont=sf]{subfig}
\usepackage{amsmath} 
\usepackage{amssymb}  
\usepackage{algorithm}
\usepackage{algorithmic}
\usepackage{booktabs}
\usepackage{multirow}
\usepackage{balance}
\usepackage{cite}
\usepackage{hyperref}
\usepackage{xcolor}

\IEEEoverridecommandlockouts                              

\makeatletter
\newcommand*\titleheader[1]{\gdef\@titleheader{#1}}
\AtBeginDocument{%
	\let\st@red@title\@title
	\def\@title{%
		\bgroup\normalfont\large\centering\@titleheader\par\egroup
		\vskip1.5em\st@red@title}
}
\makeatother

\title{\LARGE \bf
Time-Optimal Gate-Traversing Planner for Autonomous Drone Racing
}
\titleheader{\protect\textcolor{gray}{This paper has been accepted for publication at the IEEE International Conference on Robotics and Automation, Yokohama, 2024 ©IEEE}}
\author{Chao Qin$^{1}$, Maxime S.J. Michet$^{2}$, Jingxiang Chen$^{2}$, and Hugh H.-T. Liu$^{1}$
\thanks{$^{1}$Chao Qin and Hugh H.-T. Liu are with the University of Toronto Institute for Aerospace Studies (UTIAS), Toronto, Canada {\tt\small chao.qin@mail.utoronto.ca, hugh.liu@utoronto.ca}}
\thanks{$^{2}$Maxime S.J. Michet and Jingxiang Chen are with the University of Toronto Division of Engineering Science (EngSci), Toronto, Canada {\tt\small max.michet@mail.utoronto.ca, harryjx.chen@mail.utoronto.ca}}
}

\begin{document}

\maketitle
\thispagestyle{empty}
\pagestyle{empty}

\begin{abstract}
In drone racing, the time-minimum trajectory is affected by the drone's capabilities, the layout of the race track, and the configurations of the gates (e.g., their shapes and sizes). However, previous studies neglect the configuration of the gates, simply rendering drone racing a waypoint-passing task. This formulation often leads to a conservative choice of paths through the gates, as the spatial potential of the gates is not fully utilized. To address this issue, we present a time-optimal planner that can faithfully model gate constraints with various configurations and thereby generate a more time-efficient trajectory while considering the single-rotor-thrust limits. Our approach excels in computational efficiency which only takes a few seconds to compute the full state and control trajectories of the drone through tracks with dozens of different gates. Extensive simulations and experiments confirm the effectiveness of the proposed methodology, showing that the lap time can be further reduced by taking into account the gate's configuration. We validate our planner in real-world flights and demonstrate super-extreme flight trajectory through race tracks.
\end{abstract}

\section*{SUPPLEMENTARY MATERIAL}

\textbf{Code:} \href{https://github.com/FSC-Lab/TOGT-Planner}{https://github.com/FSC-Lab/TOGT-Planner}

\section{INTRODUCTION}\label{sect_introduction}
\begin{figure}[!t]
\centering
\subfloat[]{\includegraphics[width=0.25\textwidth]{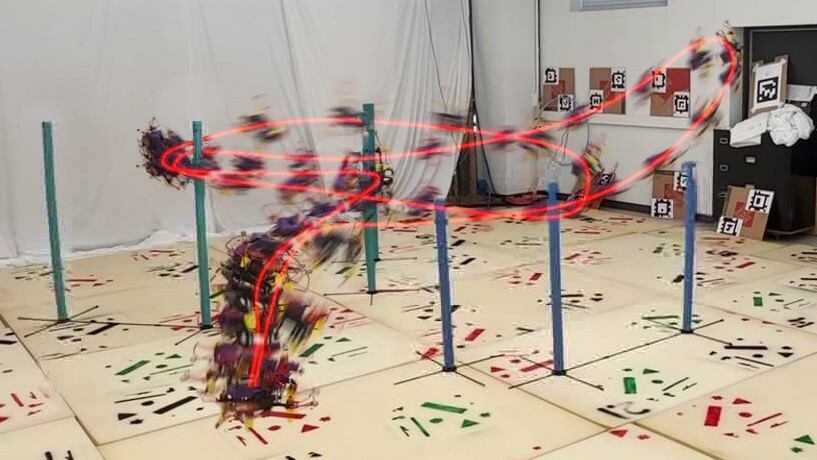}\label{fig_exp}}
\hfil
\subfloat[]{\includegraphics[width=0.20\textwidth]{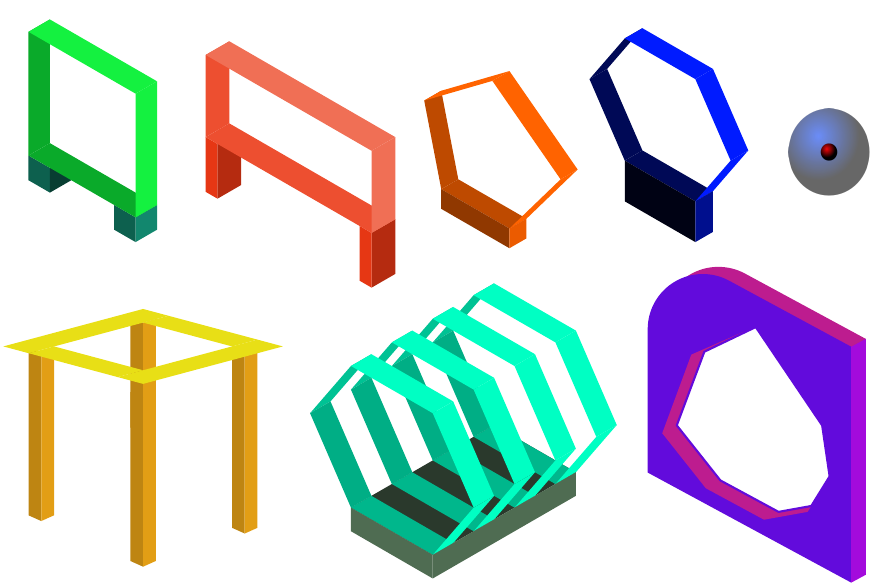}\label{fig_all_gates}}
\hfil
\subfloat[]{\includegraphics[width=0.225\textwidth]{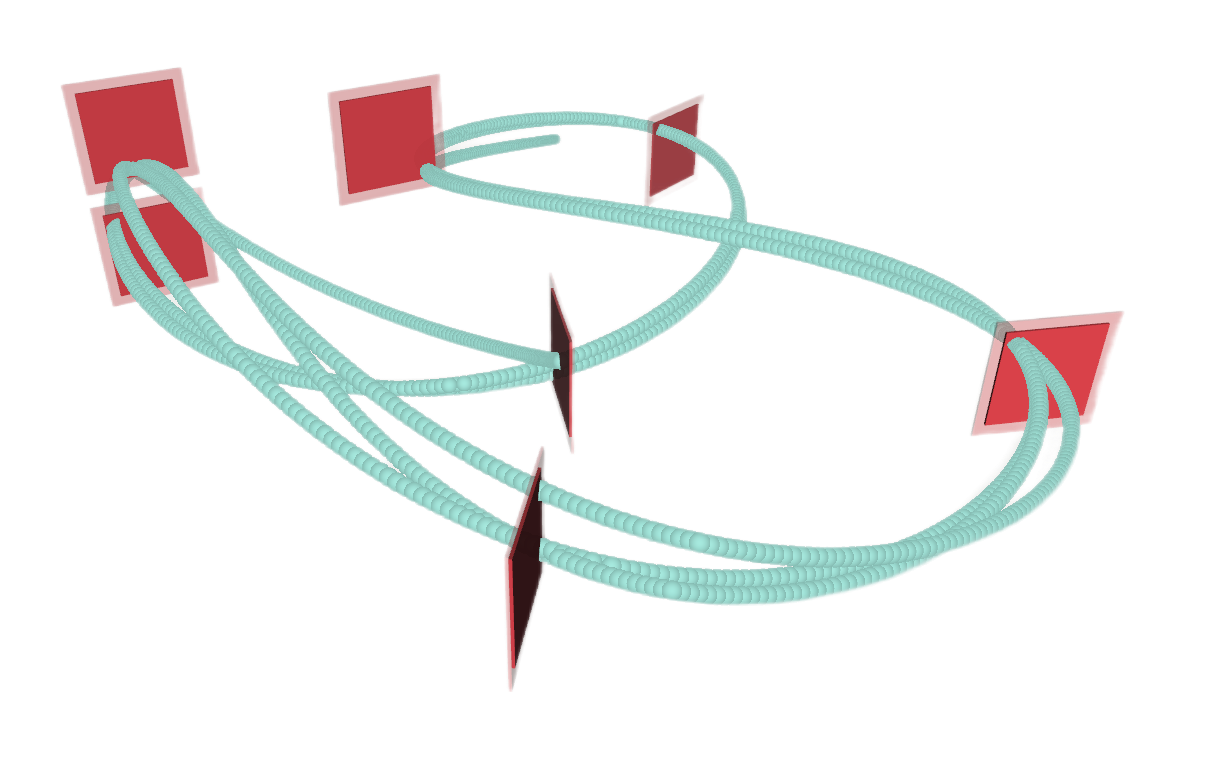}\label{fig_path_togt_front}}
\hfil
\subfloat[]{\includegraphics[width=0.225\textwidth]{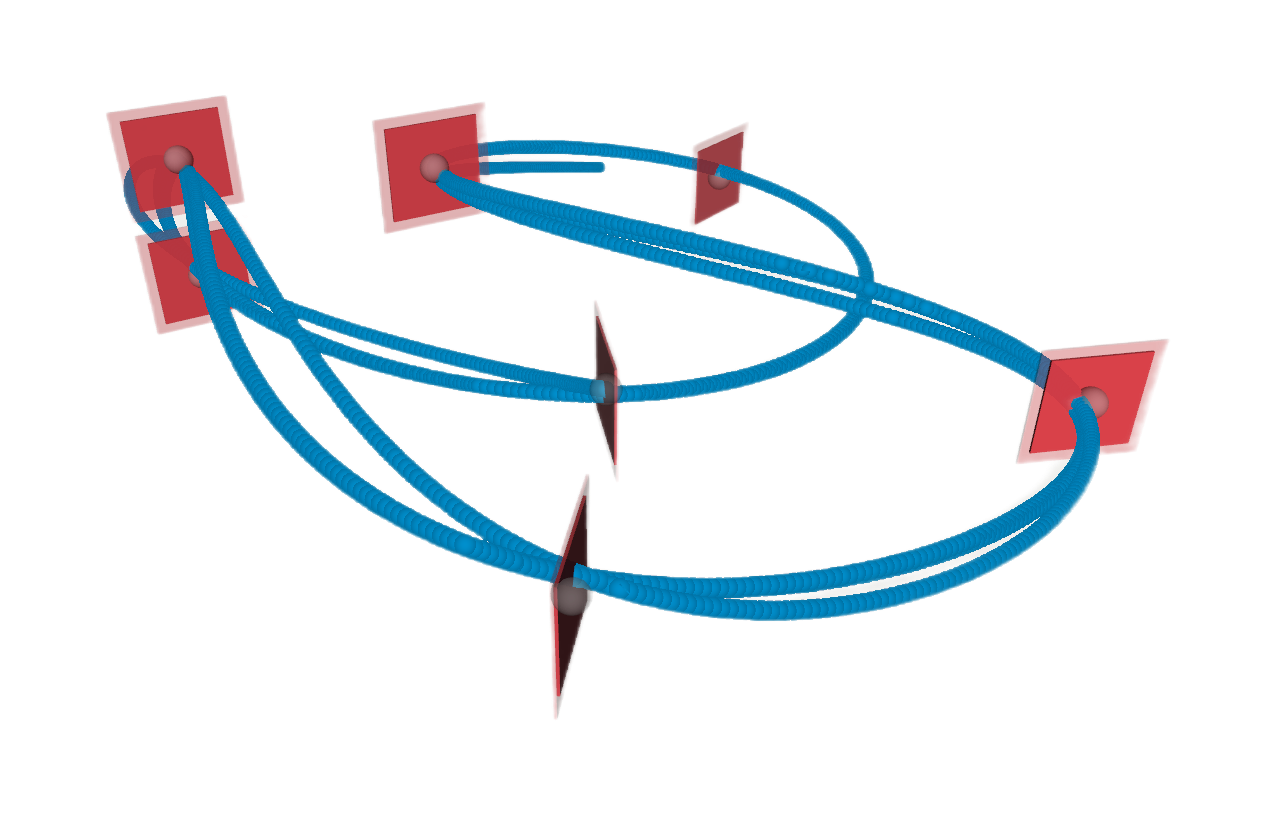}\label{fig_path_towp_front}}
\caption{(a) A time-optimal flight path generated by the proposed TOGT planner and executed in a motion capture room. (b) Our framework supports a wide range of gate shapes. (c-d) Comparison of trajectories obtained from our method with gate constraints (left figure) and with waypoint constraints (right figure). We can clearly see that the trajectory on the left is more time-efficient as it traverses the gates along their boundaries.} \label{fig_flight_demo}
\end{figure}

First-person-view (FPV) drone racing is a rapidly expanding e-sport that has attracted significant attention from the public. Fig. \ref{fig_exp} shows a typical racing scenario. In brief, the goal is to operate the quadrotor through gates in a predetermined sequence at a minimum lap time. We refer to the problem of generating such a time-minimum trajectory as the time-optimal gate-traversing (TOGT) problem. Solving the TOGT problem for general race tracks while considering the full quadrotor dynamics is very challenging. On the one hand, a single race track may contain many gate types as shown in Fig. \ref{fig_all_gates}, and the solver is required to accommodate all of them in one optimization problem. On the other hand, a substantial amount of constraints must be considered, such as the single-rotor thrust and gate constraints.

To manage computational complexity, state-of-the-art approaches \cite{foehn2021time, penicka2022minimum, zhou2023efficient, foehn2022alphapilot, romero2022time, ramos2021minimum} simplify the TOGT problem to a time-optimal waypoint-passing (TOWP) problem, where each gate's center is associated with a waypoint to guarantee successful gate traversal. While this methodology has already led to exceptional racing performance, it can only yield an approximate solution to the TOGT problem unless the gate is truly a point. It can be seen from Fig. \ref{fig_flight_demo} that the TOGT trajectory (left bottom) can always produce a shorter path and thus a faster lap time by fully utilizing the available free space within the gates. Therefore, to achieve true time optimality in racing, it is necessary to take the gate's shape and size into account. Additionally, concerns about computational efficiency arise when tackling general race tracks typically including dozens of gates. Current research mainly focuses on problems with less than 20 gates, with computation times that have already exceeded minutes or even hours \cite{foehn2021time}. Hence, their scalability still requires further investigation. To sum up, it remains an open problem to efficiently generate TOGT trajectories on general race tracks.

In this paper, we present a fast TOGT planner that can tackle dozens of gates with varying shapes in one single optimization problem within seconds. Our method supports all gate configurations listed in Fig. \ref{fig_all_gates}. Moreover, to obtain high trajectory quality, we account for the full quadrotor dynamics and constraints such as the single-rotor-thrust limit. We apply polynomials as the trajectory representation and employ the differential-flat property of the quadrotor to avoid numerical integration of the system dynamics. Despite that polynomials are inherently smooth by definition, our approach yields a close or even shorter lap time than the state-of-the-art method \cite{foehn2021time} in most tested race tracks. This is because the spatial potential of gates is fully explored as shown in Fig. \ref{fig_path_togt_front}. In addition, our planner can also address the TOWP problem by treating waypoints as special ball gates with a small radius (See Fig. \ref{fig_path_towp_front}). Our contributions are summarized below:
\begin{itemize}[]
\item We propose a comprehensive formulation of the TOGT problem and provide a lightweight solver with a second-wise computation time. 
\item To our best knowledge, this is the first approach capable of handling race tracks with such a diverse array of gate types, and the first polynomial-based method that can enforce single-rotor-thrust constraints exactly.
\item We validate the proposed approach through extensive simulation and real-world scenarios, demonstrating its great potential of generating racing benchmarks for general race tracks.
\end{itemize}

\section{Related Work}\label{sect_related_work}

Time-optimal planners for autonomous drone racing can be mainly categorized into discretization-based methods and polynomial-based methods.

Discretization-based methods employ time-discretized \cite{lai2006time, shen2023aggressive, romero2022model} or space-discretized trajectories \cite{spedicato2017minimum, arrizabalaga2022towards, van2013time} to represent the drone's state and control. Foehn et al. \cite{foehn2021time} addressed the time-allocation problem for discretized states by introducing a complementary progress constraint (CPC), leading to a tractable solution to the TOWP problem with the full quadrotor models. Penicka et al. \cite{penicka2022minimum} tackled the same problem using a sampling-based framework with hierarchical model refinement. However, these approaches may take minutes or even hours to obtain a solution. Zhou et al. \cite{zhou2023efficient} effectively reduced the computation burden by manually allocating waypoint constraints to specific states and refining the sampling period between consecutive waypoints. Romero et al. \cite{romero2022model, romero2022time} formulated the TOWP problem into a model predictive contouring control problem to enable real-time computation. Additionally, Spedicato et al. \cite{spedicato2017minimum} showed that converting the quadrotor state space into a traverse dynamics formulation along a geometric reference path offers an alternative perspective to the drone racing problem. Following this technique, Arrizabalaga et al. \cite{arrizabalaga2022towards} proposed a time-optimal planner for tunnel-like race tracks. While discretized controls can fit time-optimal control trajectories (e.g. bang-bang control \cite{hehn2012performance}) very well, this representation suffers from the curse of dimensionality, especially when the required trajectory duration is considerable.

Polynomial-based methods are widely used in generating smooth collision-free trajectories for quadrotors \cite{mellinger2012trajectory, richter2016polynomial, ji2022elastic}, but they are less popular in time-minimum missions due to their limited capacity of representing trajectories with abrupt changes. Nonetheless, their ability to simulate trajectories using a minimal number of coefficients makes them well-suited for applications where computational efficiency outweighs time optimality, such as real-time path replanning. Han et al. \cite{han2021fast} applied SE(3) constraints to ensure collision-free flight through narrow race courses and achieve a near-time-optimal solution by including time to the cost function. Wang et al. \cite{wang2023polynomial} proposed an online replanning framework to deal with dynamic gates during racing. However, these works do not aim for true time optimality as their objective functions trade-off between trajectory smoothness and time minimization. Ryou et al. \cite{ryou2021multi} present one of the few works that pursue pure time minimization using a piece-wise polynomial representation, in which Bayesian optimization is utilized to learn the optimal time allocation over polynomial segments from multi-fidelity data obtained from analytic models. However, this approach does not realistically model the true constraint of limited single-rotor thrusts, and it is difficult to extend their scheme to racing scenarios. Our approach differs from these works in that we aim for pure time-minimum trajectory in racing while considering the authentic quadrotor model and actuation limits.

In both categories, there is a notable lack of research faithfully modeling geometric constraints of the gates; works in \cite{foehn2021time, penicka2022minimum, zhou2023efficient, romero2022model, wang2023polynomial, ramos2021minimum} are indeed approximating gates by ball or waypoint constraints, while works in \cite{spedicato2017minimum, arrizabalaga2022towards, han2021fast, van2013time} model the race track as a collision-free tunnel to avoid formulating gate constraints explicitly. Although Bos et al. \cite{bos2022multi} present a system to tackle circle and rectangle gates using a multi-stage optimal control framework, it is limited to two specific shapes, lacking the flexibility to handle other shapes commonly seen in real race tracks, such as pentagons. In contrast, our algorithm can readily deal with most gate types in racing and offer users a high degree of freedom to customize gates to meet practical needs.

\section{Methodology}\label{sect_methodology}

In this section, we detail the quadrotor model and gate constraints used, formulate the time-optimal gate-traversing problem, and introduce our solution.

\subsection{Quadrotor Model}\label{subsect_quadrotor_model}
Let $\mathcal{F}^{W}$ and $\mathcal{F}^{B}$ denote the world frame and body frame, respectively. We define the state of the quadrotor as $\mathbf{x}=[\mathbf{p}^{W},\mathbf{q}_{WB},\mathbf{v}^{W},\boldsymbol{\omega}^{B}]^{T}\in\mathbb{R}^{n}$ where $n\!=\!13$, corresponding to position expressed in $\mathcal{F}^{W}$, unit quaternion rotation from $\mathcal{F}^{B}$ to $\mathcal{F}^{W}$, velocity in $\mathcal{F}^{W}$, and angular rate in $\mathcal{F}^{B}$. Let $\mathbf{\Lambda}(\mathbf{q})\in\mathbb{R}^{4\times4}$ represent the quaternion product matrix and $\mathbf{R}(\mathbf{q})\in\mathbb{R}^{3\times3}$ the rotation matrix of the corresponding quaternion. We will omit the frame indices from here on as they remain consistent throughout the description. The control includes commanded thrusts of four motors, $\mathbf{u}=[f_{1},f_{2},f_{3},f_{4}]^{T}\in\mathbb{R}^{m}$ where $m\!=\!4$. With a slight notation abuse, we also use $m$ to represent the quadrotor's mass. In addition, we use $\mathbf{J}$, $l$, and $c_{\tau}$ to denote the inertia, arm length, and torque constant of the quadrotor, respectively. The body torque is expressed as $\boldsymbol{\tau}\in \mathbb{R}^3$ and the gravity vector in $\mathcal{F}^{W}$ as $\mathbf{g}\in \mathbb{R}^3$. Now the equation of motion can be written as:
\begin{align}
\begin{array}{ll}
\dot{\mathbf{p}}=\mathbf{v}, & \dot{\mathbf{q}}=\frac{1}{2}\mathbf{\Lambda}(\mathbf{q})\left[\begin{array}{c}
0\\
\boldsymbol{\boldsymbol{\omega}}
\end{array}\right],\\
\mathbf{v}=\mathbf{g}+\frac{1}{m}\mathbf{R}(\mathbf{q})\mathbf{F}_{T}, & \dot{\boldsymbol{\omega}}=\mathbf{J}^{-1}(\boldsymbol{\tau}-\boldsymbol{\omega}\times\mathbf{J}\boldsymbol{\omega}),
\end{array}
\end{align}
where
\begin{equation}
\mathbf{F}_{T}\!=\!\left[\begin{array}{c}
0\\
0\\
\sum f_{i}
\end{array}\right],\;\boldsymbol{\tau}\!=\!\left[\begin{array}{c}
l(f_{1}\!+\!f_{2}\!-\!f_{3}\!-\!f_{4})\\
l(\!-\!f_{1}\!+\!f_{2}\!+\!f_{3}\!-\!f_{4})\\
c_{\tau}(f_{1}\!-\!f_{2}\!+\!f_{3}\!-\!f_{4})
\end{array}\right].
\end{equation}
To ensure feasible maneuvers, we impose force constraints \cite{schneider2012fault} on each motor, $f_{min}\!\leq  \! f_i \! \leq \! f_{max}$, as well as body rate constraints $|\boldsymbol{\omega}|<\boldsymbol{\omega}_{max}$. Lastly, we encapsulate all dynamic equations into $\dot{\mathbf{x}}=\mathbf{f}(\mathbf{x},\mathbf{u})$ and all state-input constraints into $\mathbf{h}(\mathbf{x},\mathbf{u})\leq\mathbf{0}$.

\subsection{Definition of Gate}\label{subsect_definition_of_gate}

Throughout this paper, a gate is defined as a three-dimensional region enclosed by the corresponding geometrical shape. Let $\mathcal{G}^{i}\subset\mathbb{R}^{3}$ denote the space of the $i$-th gate. It is considered being traversed if there exists a point $\mathbf{p}\in\mathcal{G}^{i}$ in the state trajectory. We introduce inequalities $\mathbf{h}_{\mathcal{G}}^{i}(\mathbf{p})\leq\mathbf{0}$ to indicate a successful traversal. Two base gate classes are considered below, using which all gates listed in Fig. \ref{fig_all_gates} can be properly constructed.

The first base class is called the ball gate. It can be used to specify checkpoints that the drone must pass in racing. Let $\mathbf{p}_{w}\in\mathbb{R}^{3}$ denote the center of the ball and $\delta\!\geq\!0$ its radius, which respectively correspond to the location and tolerance range of a waypoint. The enclosed space is given as: 
\begin{equation}
\mathcal{G}_{\mathcal{B}}=\{\mathbf{p}\in\mathbb{R}^{3}|\|\mathbf{p}-\mathbf{p}_{w}\|_{2}\leq\delta\},\label{equ_ball_gate}
\end{equation}

The second base class is called the convex polygon/polyhedron gate. While the polygon can be used to describe gate shapes like rectangles, pentagons, and so on, the polyhedron can be utilized to represent collision-free space of some tunnels. They share the same expression as
\begin{equation}
\mathcal{G}_{\mathcal{P}}=\{\mathbf{p}\in\mathbb{R}^{3}|\mathbf{A}\mathbf{p}\leq\mathbf{b}\},\label{equ_convex_polygon_gate}
\end{equation}
where $\mathbf{A}$ and $\mathbf{b}$ defines the half-spaces enclosing the gate. From here on, we will drop the subscripts, as the gate shape information is implied by their indices.

By concatenating multiple gates, we can create tunnels in the race track. For instance, a pentagonal tunnel can be constructed by two pentagons made by the polygon gate and one pentagonal prism made by the polyhedron gate. The order of these sub-gates can designate its entrance and exit. 

\subsection{Time-Optimal Gate-Traversing Problem} \label{sect_togt_problem}

Consider an environment with $L$ gates, denoted as $\mathcal{G}^{1},\mathcal{G}^{2},...,\mathcal{G}^{L}$, and a quadrotor at the initial state $\bar{\mathbf{x}}_{0}$. The objective of TOGT can be described as finding time-minimum trajectories passing through these gates and finally reaching the terminal state at $\bar{\mathbf{x}}_{f}$. We use $\mathbf{x}:[0,t_f]\mapsto\mathbb{R}^{n}$ and $\mathbf{u}:[0,t_f]\mapsto \mathbb{R}^{m}$ to express the corresponding state and control trajectory, respectively, where $t_f$ denotes the total time of the trajectory. The TOGT problem is given as:
\begin{subequations}
	\begin{align}
	\min_{\mathbf{x},\mathbf{u},t_{f}}\quad & t_{f}\\
	\textrm{s.t.}\quad & \mathbf{x}(0)=\bar{\mathbf{x}}_{0},\;\mathbf{x}(t_{f})=\bar{\mathbf{x}}_{f},\\
	& \dot{\mathbf{x}}=\mathbf{f}(\mathbf{x},\mathbf{u}),\;\mathbf{h}(\mathbf{x},\mathbf{u})\leq\mathbf{0},\\
	& \exists \;0<t_{1}<t_{2}<...<t_{L}<t_{f},\\
	& \mathbf{h}_{\mathcal{G}^{i}}(\mathbf{p}_{\mathbf{x}(t_{i})})\leq\mathbf{0},\;1\leq i\leq L,
	\end{align}\label{equ_cttogt}
\end{subequations}
where $t_{i}$ is the time passing the $i$-th gate and $\mathbf{p}_{\mathbf{x}(t_{i})}$ is the position member of $\mathbf{x}(t_{i})$.

\subsection{Problem Transformation} \label{sect_problem_transformation}

We can see that the inequality (\ref{equ_cttogt}e) is directly related to $\mathbf{x}(t_i)$ and consequently indirectly linked to $t_i$ and $t_f$, making it a time-dependent constraint. To ease the computation, it is beneficial to perform certain transformations to decouple the time from the gate constraints.

\begin{figure}[!h]
	\centering
	\includegraphics[width=0.4\textwidth]{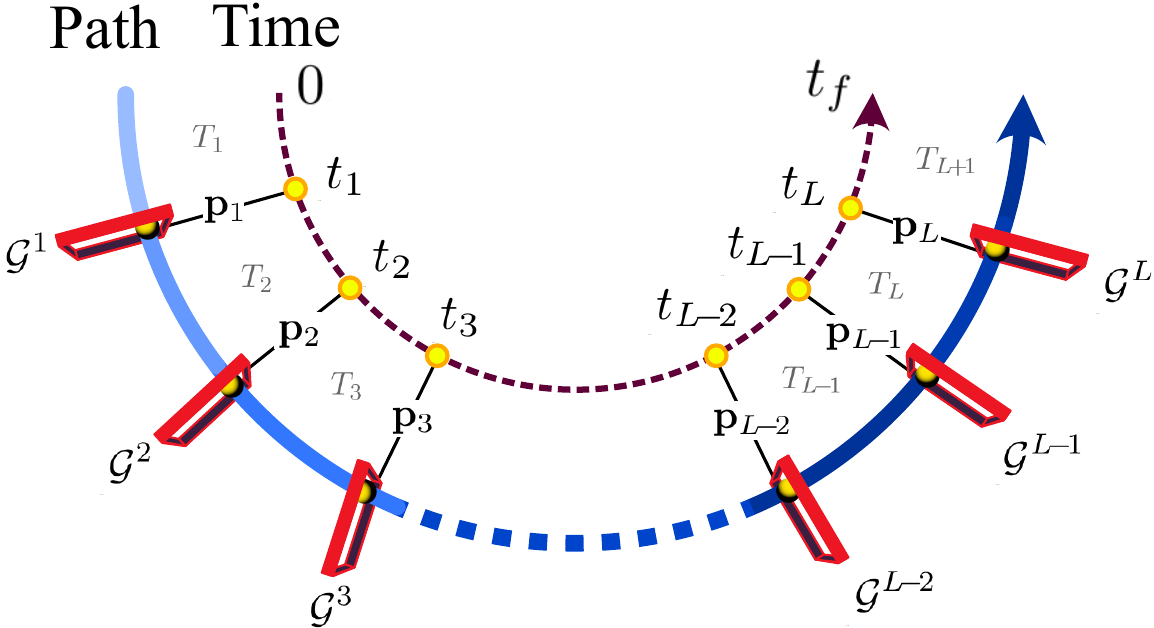}
	\caption{Illustration of the segmented trajectories based on the gate order.}
	\label{fig_togt_graph}
\end{figure}

To begin with, we segment the trajectory into $(L\!+\!1)$ pieces according to the gate order and assign each gate a waypoint, as shown in Fig. \ref{fig_togt_graph}. Let $\mathbf{P}=[\mathbf{p}_{1}^{T},\mathbf{p}_{2}^{T},...,\mathbf{p}_{L}^{T}]^{T}\in\mathbb{R}^{3L}$ denote the vector storing all resulting waypoints. The times allocated for all segments are given as $\mathbf{T}=[T_{1},T_{2},...,T_{L},T_{L\!+\!1}]^{T}\!\in\!\mathbb{R}_{>}^{L+1}$.  We can derive the new expression for the total trajectory duration as $t_{f} \triangleq T_{\Sigma} = \sum_{k=1}^{L+1} T_{k}$ and the traversal time of the $i$-th gate as $t_{i} \triangleq t_{\Sigma_{i}} = \sum_{k=1}^{i} T_{k}$.

After the waypoints are given, we put all feasible time allocation vectors for this specific waypoint flight into a set:
\begin{align}\begin{split}\mathcal{T}(\mathbf{P})=\{\mathbf{T}\in\mathbb{R}_{>}^{L+1} & \;|\;\exists\mathbf{x},\mathbf{u}\!:\![0,\!T_{\Sigma}]\!\rightarrow\!\mathbb{R}^{n}\!,\!\mathbb{R}^{m}\\
\text{ s.t. } & \mathbf{x}(0)=\bar{\mathbf{x}}_{0},\;\mathbf{x}(T_{\Sigma})=\bar{\mathbf{x}}_{f}\\
& \dot{\mathbf{x}}=\mathbf{f}(\mathbf{x},\mathbf{u}),\mathbf{h}(\mathbf{x},\mathbf{u})\leq\mathbf{0}\\
& \mathbf{p}_{\mathbf{x}(t_{\Sigma_{i}})}=\mathbf{p}_{i},\;1\leq i\leq L\}.
\end{split}\label{equ_feasible_time_alloc}
\end{align}
Note that $\mathcal{T}(\mathbf{P})$ encompasses all possible functionals to represent $\mathbf{x}$ and $\mathbf{u}$. According to this definition, if $\mathbf{T}\in\mathcal{T}(\mathbf{P})$, there must exist a dynamically feasible trajectory passing all specified waypoints at the specified times. 

Then, we can rewrite (\ref{equ_cttogt}) as
\begin{subequations}
	\begin{align}\min_{\mathbf{P},\mathbf{T}}\quad & T_{\Sigma}+I_{\mathcal{T}(\mathbf{P})}(\mathbf{T})\label{equ_cttogt_compact_a}\\
	\textrm{s.t.}\quad & \mathbf{h}_{\mathcal{G}^{i}}(\mathbf{p}_{i})\leq\mathbf{0},\;1\leq i\leq L,\label{equ_cttogt_compact_b}
	\end{align}\label{equ_cttogt_compact}
\end{subequations}
where
\begin{align}I_{\mathcal{T}(\mathbf{P})}(\mathbf{T})=\begin{cases}
\begin{array}{c}
0\\
\infty
\end{array} & \begin{array}{c}
\text{if}\;\mathbf{T}\in\mathcal{T}(\mathbf{P}),\\
\text{if}\;\mathbf{T}\notin\mathcal{T}(\mathbf{P}).
\end{array}\end{cases}\label{equ_indicator_function}
\end{align}
Checking the equivalence between (\ref{equ_cttogt}) and (\ref{equ_cttogt_compact}) is pretty straightforward and we omit the proof here.

It can be seen that the inequality (\ref{equ_cttogt_compact_b}) is no longer time-dependent, which is key for us to eliminate it in Section \ref{subsect_gate_constraint_elimination}. However, solving $\mathcal{T}(\mathbf{P})$ and subsequently $I_{\mathcal{T}(\mathbf{P})}(\mathbf{T})$ is still intractable, because we cannot travel the entire functional space due to the curse of dimentionality \cite{diehl2011numerical}. Our strategy is to focus on one specific functional and aim for an approximate solution, $I_{\mathcal{\hat{T}}(\mathbf{P})}(\mathbf{T})\approx I_{\mathcal{T}(\mathbf{P})}(\mathbf{T})$. In the next section, we will discuss how to select a suitable functional.


\subsection{Functional Selection and Calculation of $I_{\mathcal{\hat{T}}(\mathbf{P})}(\mathbf{T})$}\label{subsect_functional_selection}

The most suitable functional varies with the design objective. If the goal is true time optimality like in \cite{foehn2021time}, the piece-wise constant control becomes a viable candidate, albeit with the drawback of large computation costs. In our case, to enable rapid trajectory generation, we prioritize computational efficiency over the utmost time optimality. Therefore, we prefer a functional that (i) can eliminate as many constraints in (\ref{equ_feasible_time_alloc}) as possible and (ii) has a minimal number of coefficients, using which the problem complexity can be greatly alleviated. 

The minimum-control (MINCO) trajectory functional proposed in \cite{wang2022geometrically} meets the above criteria very well. This functional in essence is a set of piece-wise polynomials that minimize energies (e.g., jerk). However, we apply it primarily as a lightweight tool to compute polynomial coefficients, rather than taking energy minimization as a design objective like in \cite{ji2022elastic, han2021fast, wang2023polynomial}. By parameterizing the quadrotor's flat output trajectory using piece-wise polynomials obtained from this method, we not only can eliminate constraints on system dynamics, initial state, and terminal state but also ensure precise waypoint traversal. These properties enable it to remove a large number of constraints from (\ref{equ_feasible_time_alloc}), leaving only the state-input constraints. Another advantage of this functional is that solving the polynomial coefficients is as straightforward as solving a matrix equation, making the trajectory generation and evaluation very efficient.

Let $\mathbf{y}\!:\![0,t_{f}]\!\rightarrow\!\mathbb{R}^{m}$ denote the flat output trajectory which is implemented as piece-wise polynomials with an order of $s\!=\!5$. We have $\mathbf{y}(t)\!=\![\mathbf{p}^T,\psi]^T$ where $\psi$ is the yaw angle \cite{faessler2017differential}. The state and control trajectories, along with their associated constraints, can now be expressed as functions of $\mathbf{y}$ and 
its derivatives up to the $s$-th order:
\begin{align}
\mathbf{x} & =\Psi_{\mathbf{x}}(\mathbf{y},\mathbf{\dot{y}},...,\mathbf{y}^{(s-1)})\triangleq\Psi_{\mathbf{x}}(\mathbf{y}^{[s-1]}),\label{equ_flat_state}\\
\mathbf{u} & =\Psi_{\mathbf{u}}(\mathbf{y},\mathbf{\dot{y}},...,\mathbf{y}^{(s)})\triangleq\Psi_{\mathbf{u}}(\mathbf{y}^{[s]}),\label{equ_flat_control}\\
\mathbf{h}&(\mathbf{x},\mathbf{u})=\mathbf{h}_{\Psi}(\mathbf{y}^{[s]})\leq\mathbf{0},\label{equ_flat_state_input_constraint}
\end{align}
where $\Psi_{\mathbf{x}}:\mathbb{R}^{m\times(s\!-\!1)}\!\rightarrow\!\mathbb{R}^{n}$ and $\Psi_{\mathbf{u}}:\mathbb{R}^{m\times s}\!\rightarrow\!\mathbb{R}^{n}$ are defined in \cite{mellinger2011minimum, faessler2017differential, ferrin2011differential}. Since (\ref{equ_flat_state_input_constraint}) is the only remaining constraint in (\ref{equ_feasible_time_alloc}), any violation at an arbitrary time would yield $\mathbf{T} \!\notin \! \mathcal{\hat{T}}(\mathbf{P})$. Now we define $I_{\mathcal{\hat{T}}(\mathbf{P})}(\mathbf{T})$ as
\begin{align}\begin{split} & I_{\mathcal{\hat{T}}(\mathbf{P})}(\mathbf{T})\!=\!\int_{0}^{T_{\Sigma}}\!\max[\mathbf{h}_{\Psi}(\mathbf{y}^{[s]}(t)),\!\mathbf{0}]^{3}dt,\\
& \!\approx\!\sum_{i=1}^{L\!+\!1}\sum_{j=0}^{\kappa_{i}}\max[\mathbf{h}_{\Psi}(\mathbf{y}^{[s]}(t_{i\!-\!1}+j\Delta t_{i})),\!\mathbf{0}]^{3}\Delta t_{i},
\end{split}\label{equ_penality}
\end{align}
where $\kappa_{i}$ the sample resolution, $t_0=0$, and $\Delta t_{i}=T_{i}/\kappa_{i}$. It can be observed that $I_{\mathcal{\hat{T}}(\mathbf{P})}(\mathbf{T})$ returns zero if all constraints are met and nonzero otherwise, functioning just like $I_{\mathcal{T}(\mathbf{P})}(\mathbf{T})$. Therefore, when $I_{\mathcal{\hat{T}}(\mathbf{P})}(\mathbf{T})$ is small, we consider the state-input constraints to be satisfied.

\subsection{Gate and Time Constraints Elimination}\label{subsect_gate_constraint_elimination}

The gate constraints in (\ref{equ_cttogt_compact_b}) can become high-dimensional when dealing with intricate gate shapes or a substantial number of gates. The inherent positivity constraints in $\mathbf{T}$ also bring an additional $L+1$ constraints to the system. These constraints not only introduce a large number of Lagrange multipliers into the solver but also contribute to slow convergence \cite{wright1999numerical}. Therefore, it is worthwhile to seek ways to eliminate these constraints. We employ the change-of-variable technique proposed in \cite{wang2022geometrically} for this purpose.

We know from \cite{wang2022geometrically} that for an arbitrary ball gate, there exist a smooth surjection $\mathbf{g}_{B}(\cdot):\mathbb{R}^{4}\rightarrow\mathcal{G}_{\mathcal{B}}$:
\begin{equation}
\mathbf{g}_{B}(\mathbf{d})=\mathbf{p}_{w}+\left[\frac{2\delta\mathbf{d}}{\mathbf{d}^{T}\mathbf{d}+1}\right]_{3},
\end{equation}
such that optimization over the new variables $\mathbf{d}$ implicitly satisfies the constraint $\mathcal{G}_{\mathcal{B}}$. Note that $\left[\cdot\right]_{v}$ returns the first $v$ entries of the input vector and here $v=3$. Similarly, for a convex polygon/polyhedron gate with $v$ corners, there also exist a smooth surjection: $\mathbf{g}_{\mathcal{P}}(\cdot):\mathbb{R}^{v}\rightarrow\mathcal{G}_{\mathcal{P}}$
\begin{equation}
\mathbf{g}_{\mathcal{P}}(\mathbf{d})=\mathbf{o}+\mathbf{V}\left[\frac{[\mathbf{d}]^{2}}{(\mathbf{d}^{T}\mathbf{d})^{2}}\right]_{v},
\end{equation}
that can serve the same purpose, where $\mathbf{o}\in\mathbb{R}^3$ is the origin of the barycentric coordinate \cite{warren2007barycentric} of the gate and $\mathbf{V}\in\mathbb{R}^{3\times v}$ its basis-vector matrix. By employing a set of new variables, $\mathbf{D}=[\mathbf{d}_{1}^{T},\mathbf{d}_{2}^{T},...,\mathbf{d}_{L}^{T}]^{T}$, as the underlying implementation of the waypoints, we ensure that these waypoints can always stay within their corresponding gates. We adopt a similar process to eliminate constraints on $\mathbf{T}$, which yields a set of new variables, $\mathbf{K} \!\in \!\mathbb{R}^{L+1}$. Now the waypoint and time allocation are obtained from $\mathbf{P}(\mathbf{D})$ and $\mathbf{T}(\mathbf{K})$, respectively.

\subsection{Unconstrained Optimization}

Utilizing new decision variables, we arrive at an unconstrained optimization problem below:
\begin{equation}
\min_{\mathbf{D},\mathbf{K}}\quad T_{\Sigma}(\mathbf{K})+I_{\mathcal{\hat{T}}(\mathbf{P}(\mathbf{D}))}(\mathbf{T}(\mathbf{K})).\label{equ_cttogt_di}
\end{equation}
We derive the analytical gradient of (\ref{equ_cttogt_di}) and solve it by using the L-BFGS algorithm \cite{liu1989limited}. After obtaining $\mathbf{P}$ from $\mathbf{D}$ and $\mathbf{T}$ from $\mathbf{K}$, we first compute the coefficients of the MINCO functional, followed by calculating the state and control trajectories via (\ref{equ_flat_state}) and (\ref{equ_flat_control}).

\begin{table}[!htbp]
	\tabcolsep=0.08cm
	\centering
	\caption{Quadrotor Parameters}\label{tab_params}
	\begin{tabular}{@{}ccccccc@{}}
		\toprule
		& $m$ [kg] & $l$ [m] & $\mathbf{J}_{diag}$ [gm$^2$] & $f_{max}$ [N] & $c_{\tau}$ [1]   & $\boldsymbol{\omega}_{max}$ [rad$\,$s$^{-\!1}$] \\ \midrule
		QuadA & 0.85  & 0.15  & [1,1,1.7]     & 6.88       & 0.05       & [15, 15, 3]             \\
		QuadB & 1.05 & 0.125 & [2.5,2.1,4.3] & 6.375     & 0.022      & [8, 8, 3]              \\ \bottomrule
	\end{tabular}
\end{table}

\section{Results} \label{sect_result}

In this section, we evaluate the performance of the proposed planner in two separate modes, gate-traversing mode (referred to as TOGT) and waypoint-passing mode (abbreviated as TOGT-WP). We compare them to the state-of-the-art approach, CPC \cite{foehn2021time}, which employs the same system model and constraint as our approach. The computational unit is an Intel Core i7-12650H CPU, and an optimization attempt is considered unsuccessful if it fails to converge within 8 hours.

\subsection{Solution Quality, Computation Time, and Scalability}\label{subsect_simulation}

\begin{figure}[!t]
	\centering
	\includegraphics[width=0.35\textwidth]{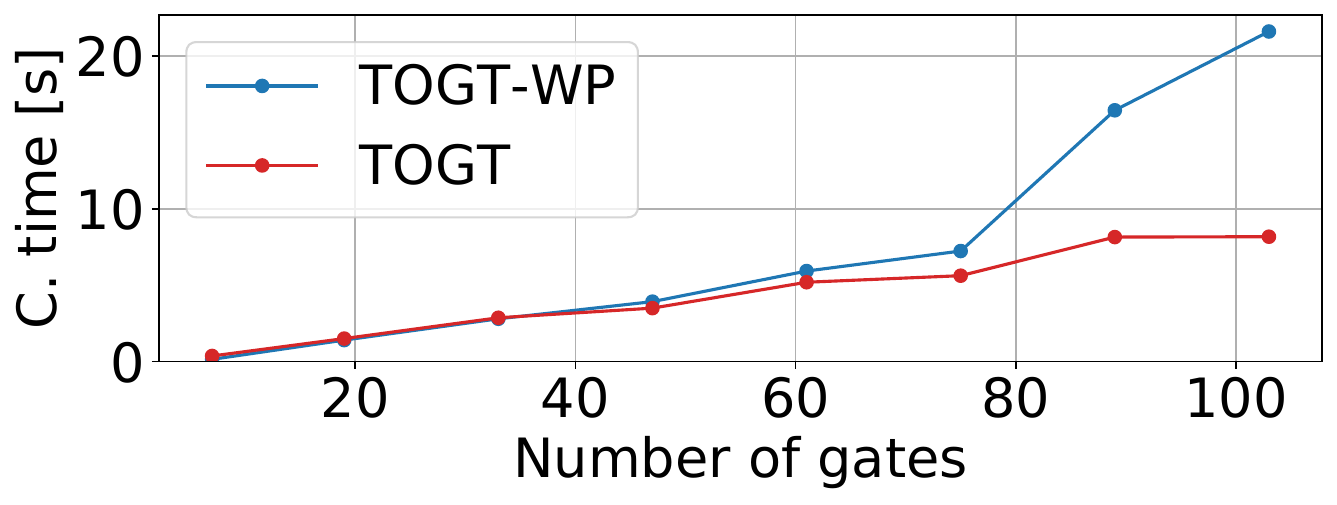}
	\caption{Our planner presents a nearly linear increase in computation time as the number of gates increases.}
	\label{fig_ctime}
\end{figure}

We simulate a racing environment described in \cite{sun2022comparative} which included seven square gates with a side length of 2.4 m. In the TOGT setup, a safe margin of 0.3 m is considered to avoid the collision, leading to an actual side length of 2.1 m. To ensure that the underlying problems for the CPC and TOGT-WP are identical, we utilize the same waypoints, which are set as the centers of the gates, and employ the same tolerance range of 0.3 m. The quadrotor parameters are provided in the QuadA configuration in Tab. \ref{tab_params}. Both the initial and terminal states are set as hovering with zero velocity. We generate trajectories with varying laps by concatenating the gates for a single lap multiple times and solving the full multi-lap problem in a single optimization. 

\begin{figure}[!t]
	\centering
	\includegraphics[width=0.45\textwidth]{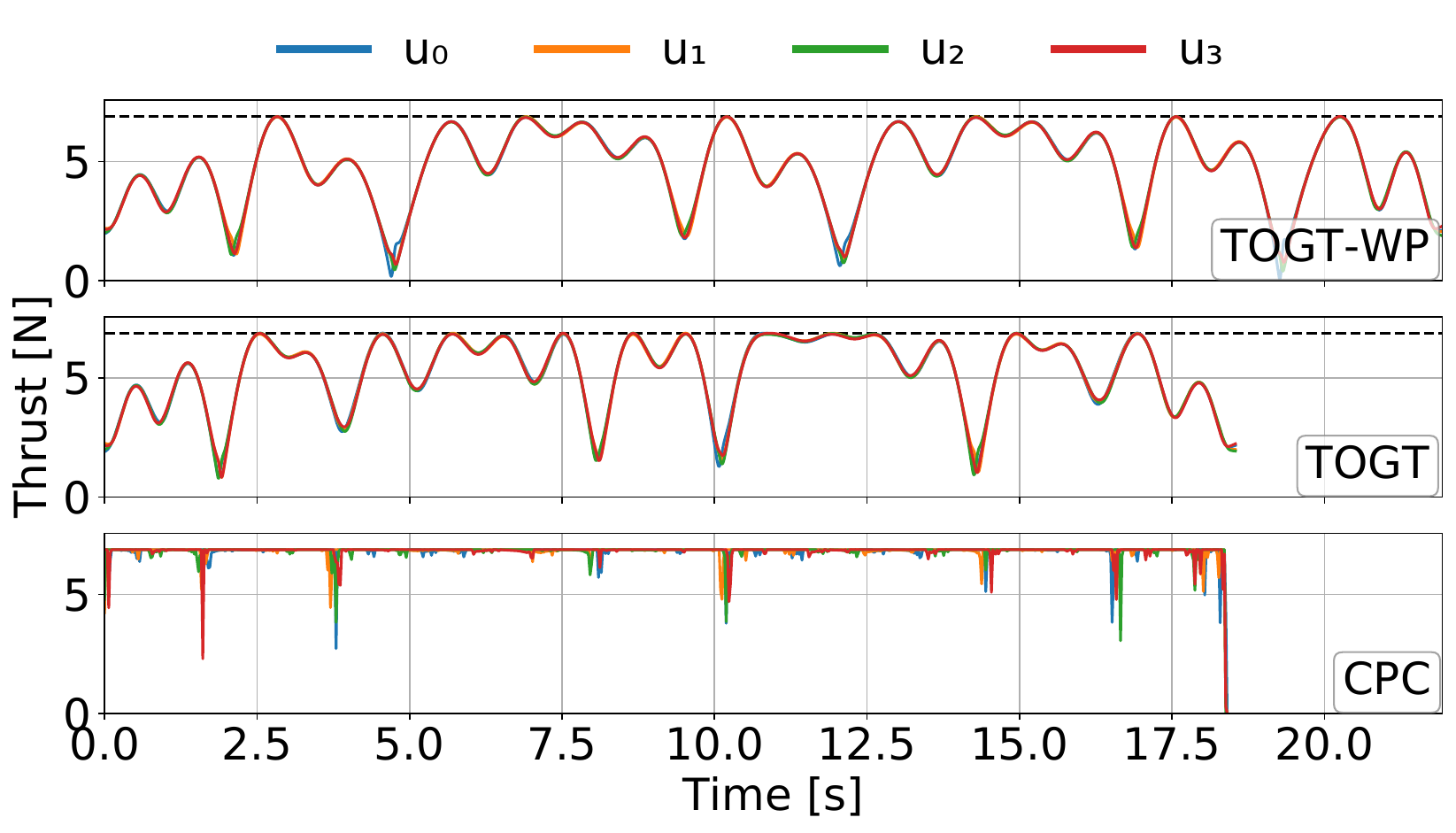}
	\caption{Comparison of single-rotor thrust trajectories from our method (the TOGT and TOGT-WP) and CPC. Although the TOGT suffers from inherent smoothness, leveraging the spatial potential of gates allows it to produce almost the same flight time as in the CPC.}
	\label{fig_compare_thrust}
\end{figure}

\begin{table}[!h]
	\tabcolsep=0.11cm
	\centering
	\caption{Comparison of computation times and trajectory durations}\label{tab_comparison}
	\begin{tabular}{p{1.0cm}cccccccc}
		\toprule
		\multirow{2}{*}{G. Num.} & \multicolumn{2}{c}{CPC\cite{foehn2021time}}         & \multicolumn{1}{l}{} & \multicolumn{2}{c}{\textbf{TOGT-WP}}                     & \multicolumn{1}{l}{} & \multicolumn{2}{c}{\textbf{TOGT}} \\ \cline{2-3} \cline{5-6} \cline{8-9} 
		& c. time {[}s{]} & $t_f$ {[}s{]} & \multicolumn{1}{l}{} & c. time {[}s{]} & \multicolumn{1}{l}{$t_f$ {[}s{]}} & \multicolumn{1}{l}{} & c. time {[}s{]} & $t_f$ {[}s{]} \\ \toprule
		\multicolumn{1}{c|}{7}     & 466              & 6.85          &                      & \textbf{0.14}            & 8.45                              &                      & \textbf{0.36}            & 7.53          \\
		\multicolumn{1}{c|}{19}    & 2718            & 18.41         &                      & \textbf{1.39}            & 21.93                             &                      & \textbf{1.50}            & 18.54         \\
		\multicolumn{1}{c|}{33}    & 9428            & 31.22         &                      & \textbf{2.79}            & 36.71                             &                      & \textbf{2.86}            & \textbf{30.87}         \\
		\multicolumn{1}{c|}{47}    & 28405              & 44.01          &                      & \textbf{3.91}            & 51.51                             &                      & \textbf{3.49}            & \textbf{43.25}         \\
		\multicolumn{1}{c|}{61}    & --            & --         &                      & \textbf{5.91}            & 66.30                             &                      & \textbf{5.18}            & \textbf{55.61}         \\
		\multicolumn{1}{c|}{75}    & --            & --         &                      & \textbf{7.22}            & 81.08                             &                      & \textbf{5.61}            & \textbf{68.00}         \\ \bottomrule
	\end{tabular}
\end{table}

Tab. \ref{tab_comparison} compares the timings and computation times of different methods with laps from 1 (7 gates) to 10.5 (75 gates). Notably, our planner is significantly faster than the CPC. While the CPC takes 7.9 hours in the task of 43 gates, our method completes the computation within 4 s. As the number of gates grows, the limitations of the CPC become increasingly evident, eventually causing its first failure in the task of 55 gates. In contrast, our method exhibits high efficiency and numerical stability across all tasks. From Fig. \ref{fig_ctime}, we observe that the computation times increase nearly linearly as the number of gates increases. 

\begin{figure}[!t]
	\centering
	\includegraphics[width=0.48\textwidth]{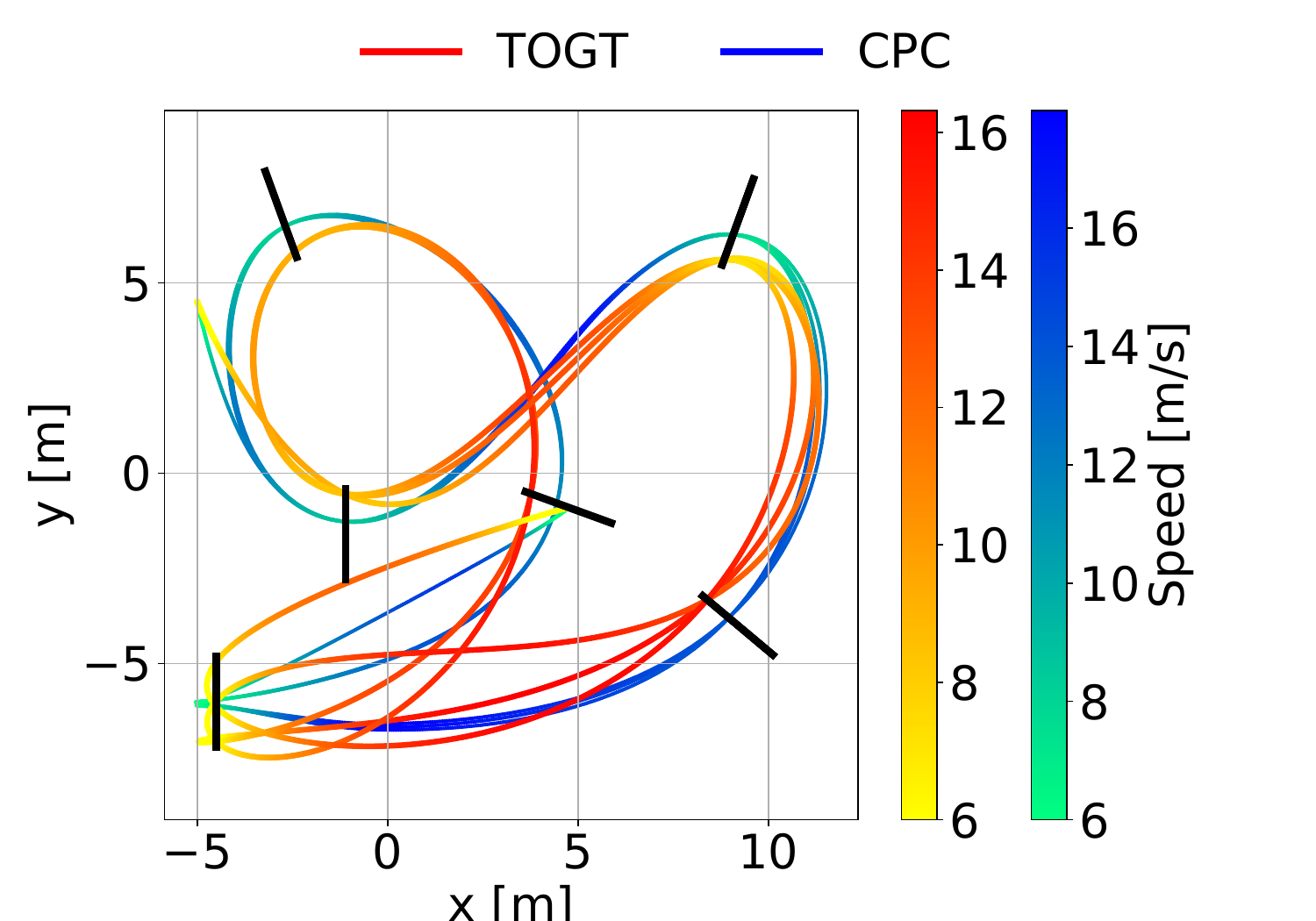}
	\caption{Comparison of trajectories obtained from the TOGT and CPC in the task of traversing 19 square gates located at 7 separate locations. Note that the gate in the lower-left corner comprises two vertically stacked gates. The trajectories are colored based on their speed profiles.}\label{fig_compare}
\end{figure}

We found that the CPC consistently outperforms the TOGT-WP in terms of solution quality, with lap times 14.6\% to 18.9\% shorter. The reason is revealed in Fig. \ref{fig_compare_thrust}, where the CPC always maintains a maximum thrust output, whereas the TOGT-WP only occasionally reaches these limits. However, this disadvantage is mitigated or even reversed when we turn to the TOGT. From Tab. \ref{tab_comparison}, we observe a notable reduction in lap times compared to the TOGT-WP. In tasks with more than 31 gates, the TOGT even begins to outperform the CPC, with lap times approximately 1\% shorter. Fig. \ref{fig_compare} demonstrates that the TOGT can keep pushing the flight aggressiveness until it reaches either the gate boundary or the drone's actuation limit. This capability results in a path that is 16.1 m shorter and a flight time that is 0.35 s shorter compared to the CPC in the task with 19 gates, despite its limited ability to maintain maximum thrust outputs.

\begin{figure}[!htbp]
	\centering
	\includegraphics[width=0.45\textwidth]{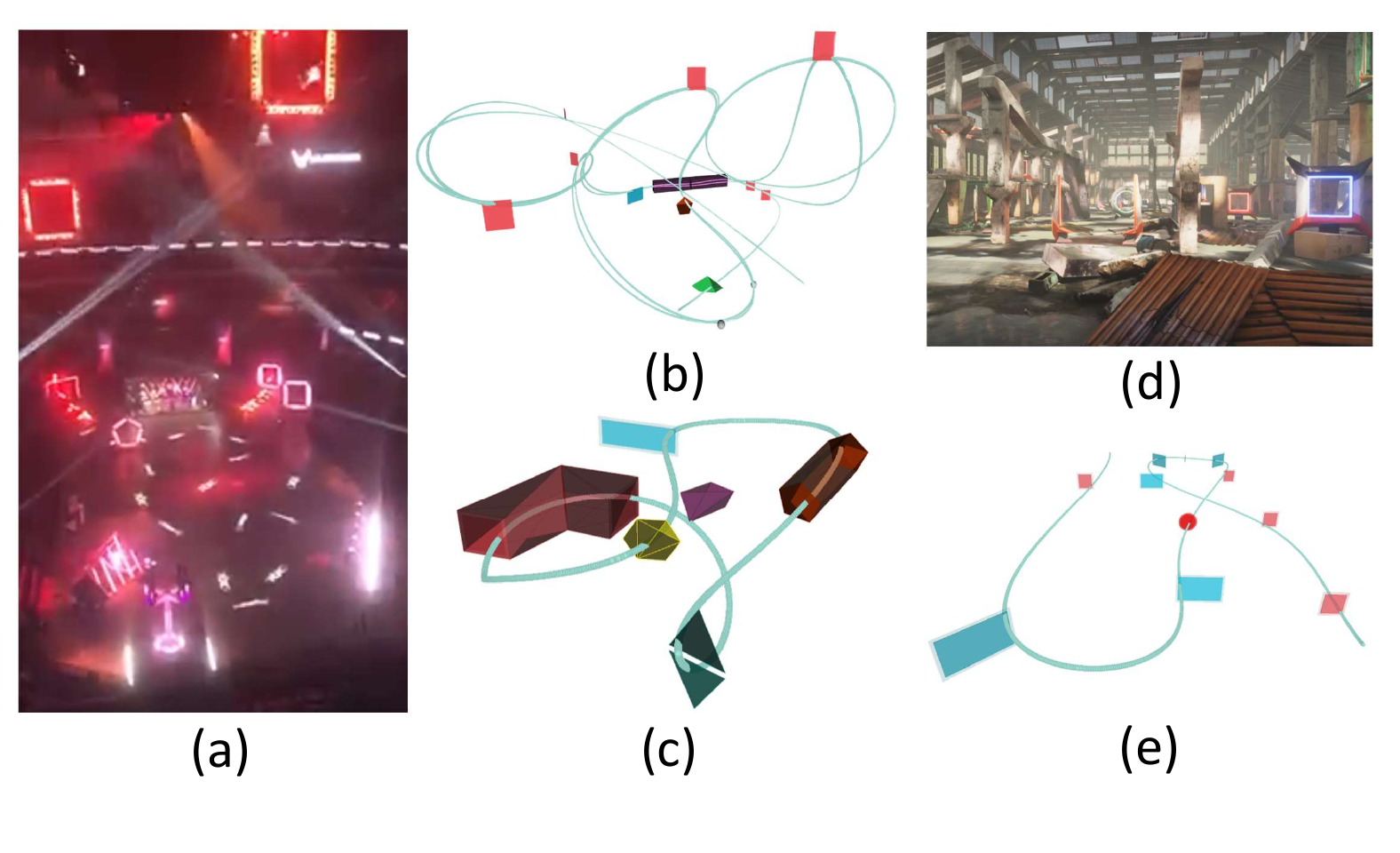}
	\caption{Trajectories generated by our method for general race tracks: (a-b) the FedExForum race track; (c) a complex race track with 6 different gates; (d-e) the FlightGoggle race track.}\label{fig_application}
\end{figure}

\subsection{Application to General Race Tracks}\label{subsect_complex_race_course}

To demonstrate its generalizability, we apply the TOGT to various race tracks, including the FedExForum track from the 2021-22 DRL World Championship, the FlightGoggle race track from the 2019 AlphaPilot Challenge \cite{foehn2022alphapilot}, and an imaginary race track showing how flexible our method can be. The generated paths are given in Fig. \ref{fig_application}, Note that the setup remains the same as in the last section.

\subsection{Experiments}

\begin{figure}[!htbp]
	\centering
	\includegraphics[width=0.35\textwidth]{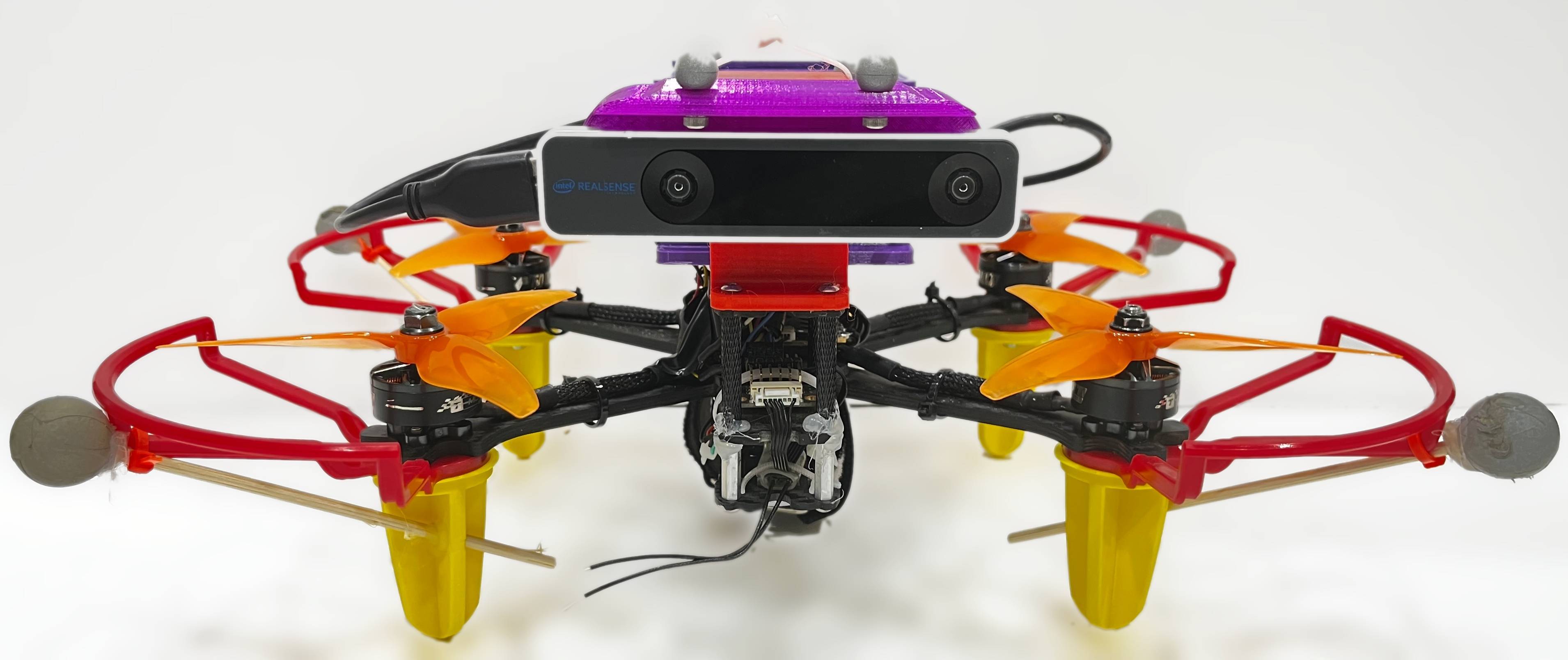}
	\caption{Platform used for the real-world experiments equipped with a Jetson Xavier computer, an Intel Realsense T265 tracking camera, a PixRacer flight controller, and infrared-reflective markers for	motion capture.}
	\label{fig_exp_drone}
\end{figure}

The labortory experimentation is performed in a motion capture room with a $4\!\times\!4\!\times\!2$ m$^3$ tracking volume. Our hardware configuration is shown in Fig. \ref{fig_exp_drone}. The onboard computer runs the \textit{Agilicious} autopilot system \cite{foehn2022agilicious}, which includes an extended Kalman filter that fuses VICON data at 100 Hz and visual-inertial odometry at 200 Hz to obtain full-state estimates, along with a model predictive controller (MPC) operating at 100 Hz. Detailed controller setups are provided in \cite{sun2022comparative}. It is worth noting that the PixRacer controller can only accept body rates and collective thrust as the lowest-level control inputs. Therefore, following \cite{foehn2021time, romero2022model}, we extract the body rates from the MPC states and compute the collective thrust by using the control commands. To maintain controllability under disturbances, we generate the trajectory with a slightly lower thrust bound than what the platform can deliver, as depicted in the QuadB configuration.

\begin{figure}[!t]
	\centering
	\subfloat[]{\includegraphics[width=0.15\textwidth]{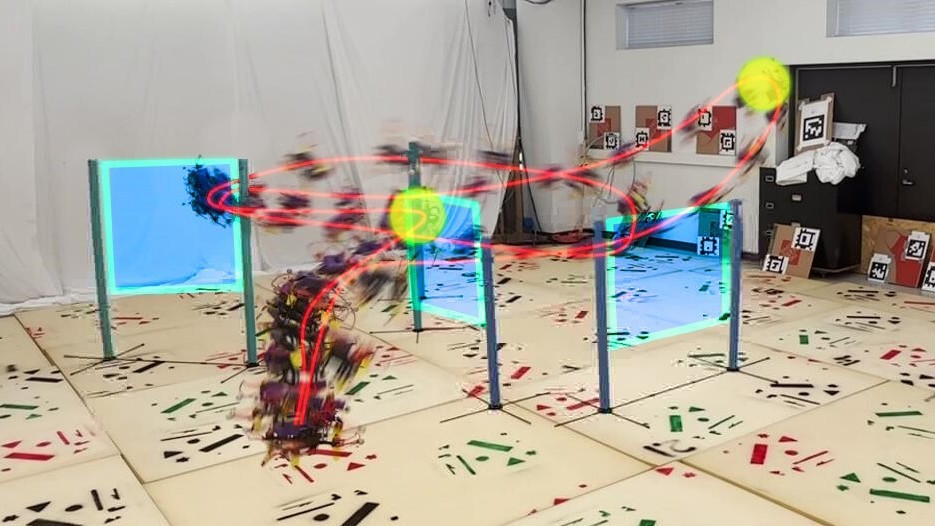}\label{fig_exp_track}}
	\hfil
	\subfloat[]{\includegraphics[width=0.15\textwidth]{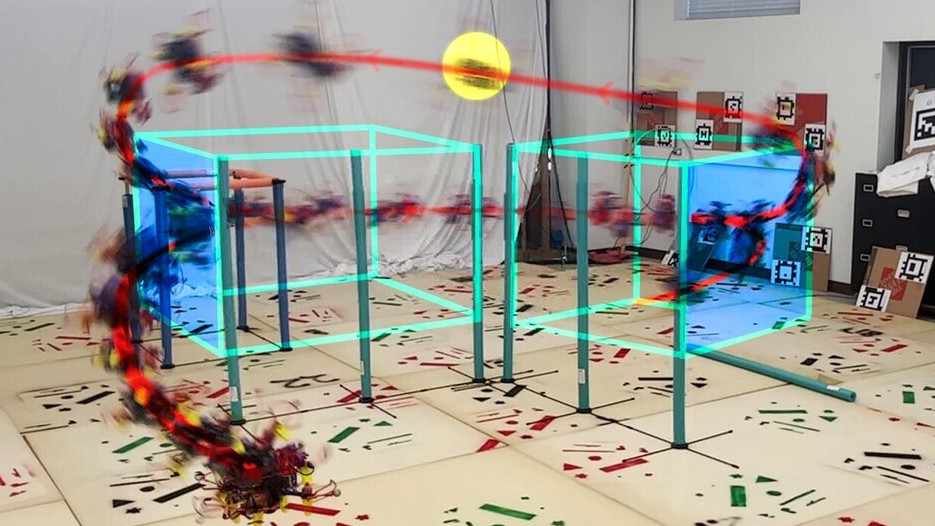}\label{fig_exp_tunnel}}
	\hfil
	\subfloat[]{\includegraphics[width=0.15\textwidth]{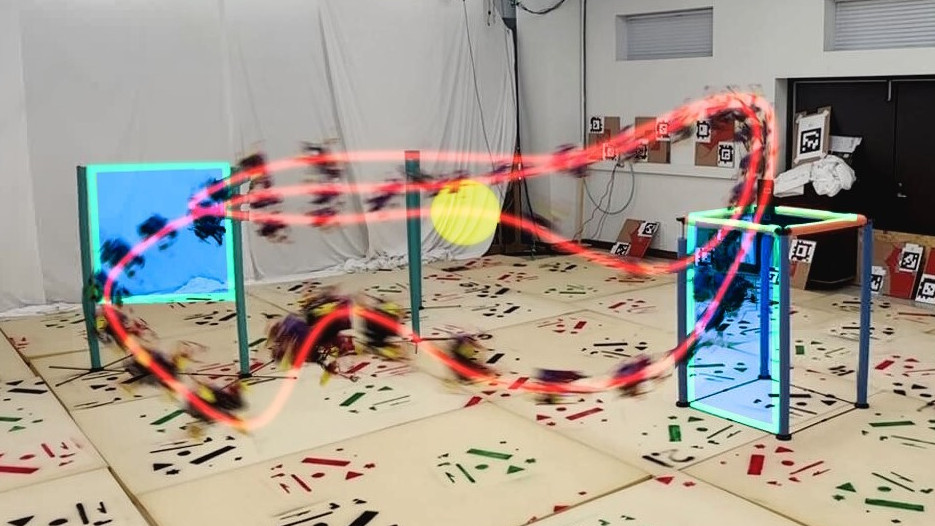}\label{fig_exp_dive}}
	\hfil
	\subfloat[]{\includegraphics[width=0.16\textwidth]{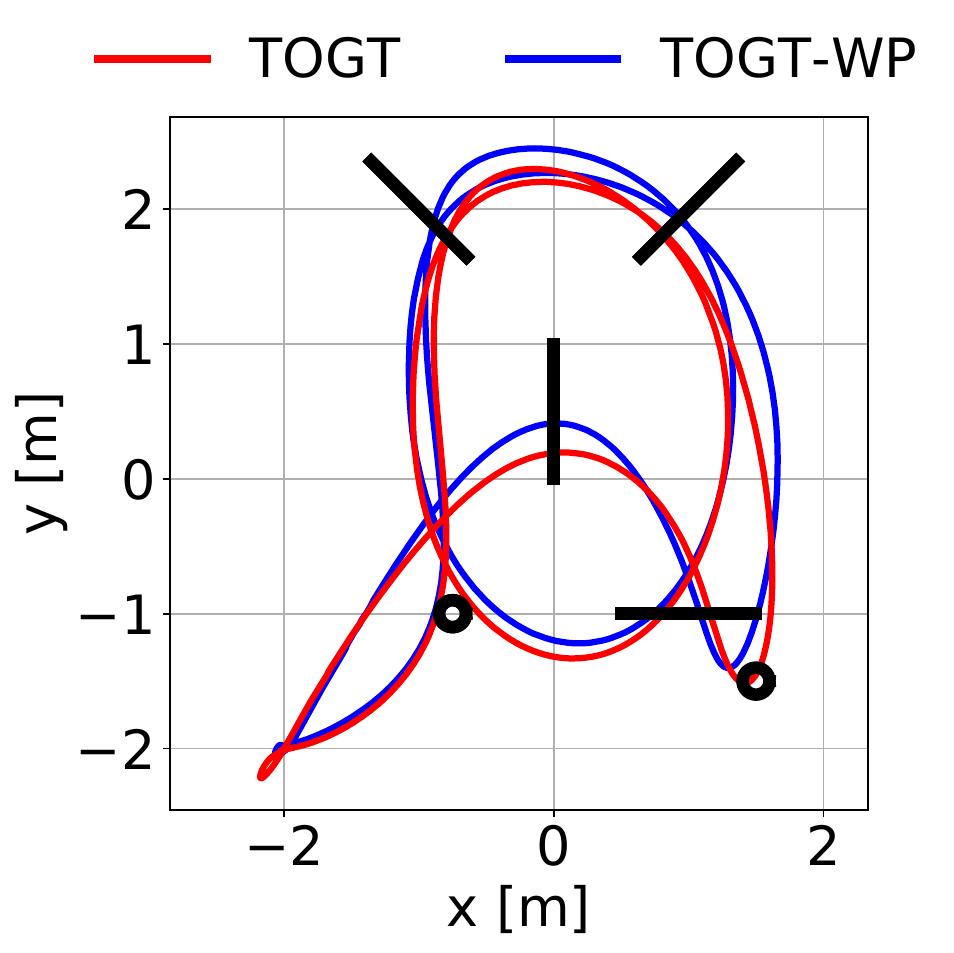}\label{fig_exp_track_data}}
	\hfil
	\subfloat[]{\includegraphics[width=0.15\textwidth]{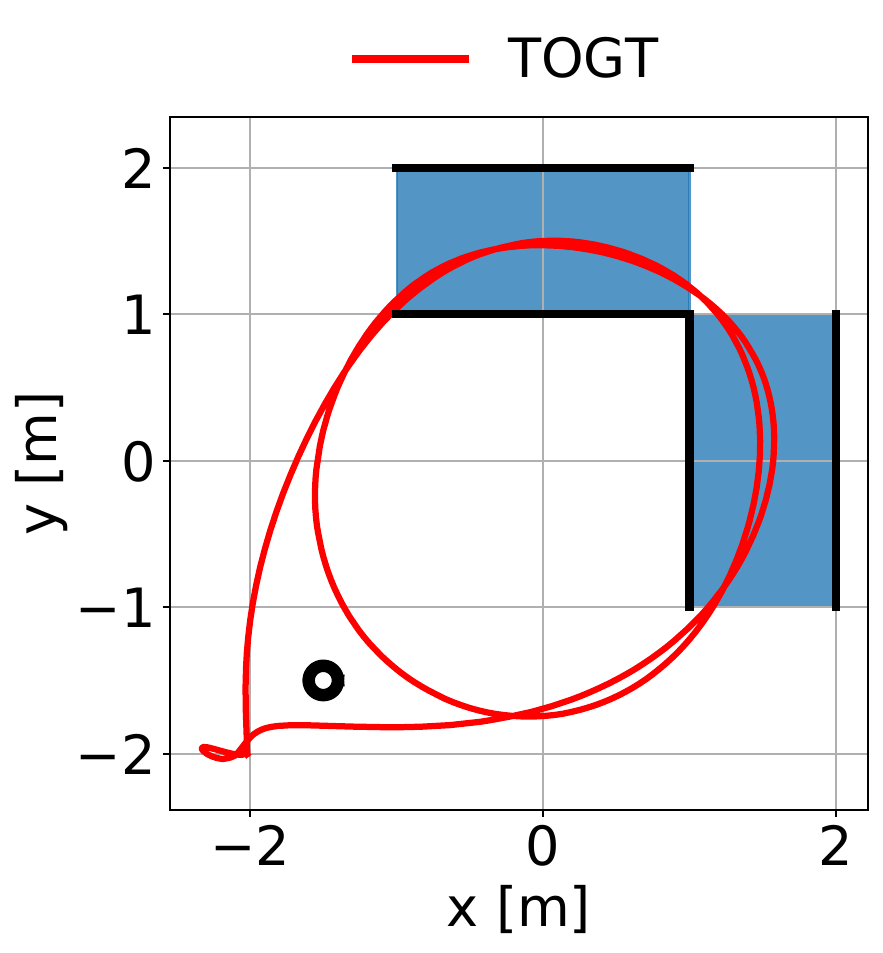}\label{fig_exp_tunnel_data}}
	\hfil
	\subfloat[]{\includegraphics[width=0.17\textwidth]{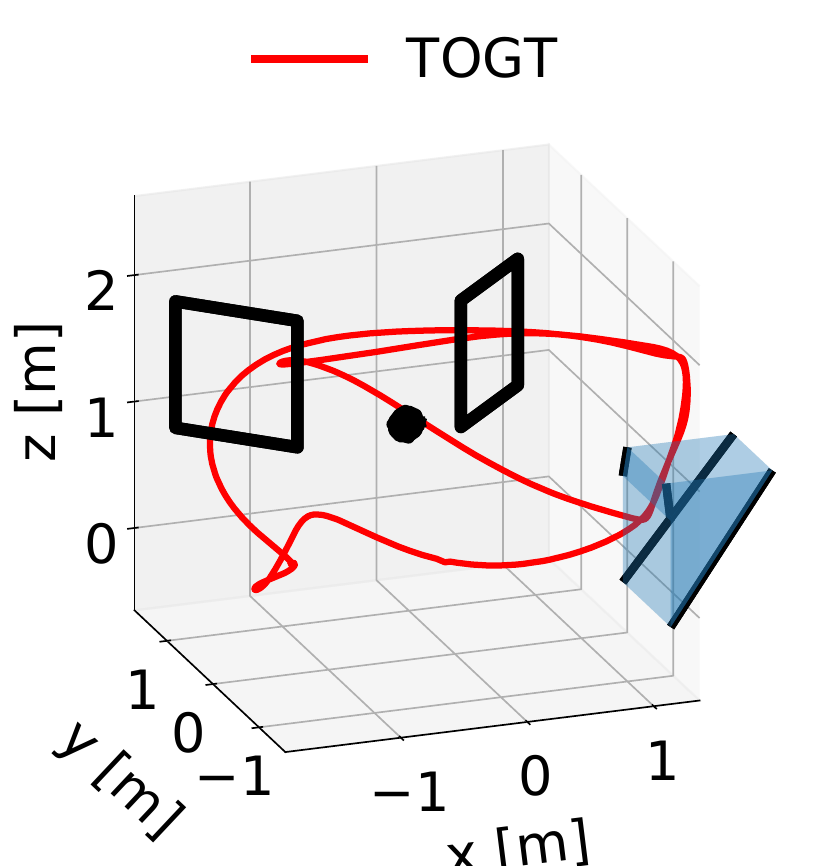}\label{fig_exp_dive_data}}
	\caption{Real-world flight trajectories in: (a,d) a course with  4 square gates; (b,e) a course with 2 tunnels; and (c,f) a course with a dive gate.}\label{fig_exp_last}
\end{figure}

We first compare the time optimality between the TOGT and TOGT-WP. The race track is illustrated in Fig. \ref{fig_exp_track} which contains a total of 10 gates, including 7 square gates with a side length of 1.0 m and 3 ball gates with a radius of 0.1 m. A margin of 0.6 m is set for each square gate to account for the actual drone size ($\approx$0.34 m) and the frame width ($\approx$0.26 m). In the TOGT-WP setup, a tolerance range of 0.1 m is set for each waypoint. The trajectories from both methods are provided in Fig. \ref{fig_exp_track_data}. The results suggest that the TOGT can generate more aggressive racing behavior in terms of space utilization. The required lap time is also shorten from 6.14 s to 5.96 s without causing any collision to gates.

In the second experiment, we evaluate the racing performance on a narrow race track made by 4 identical square tunnels and 1 ball gate. The size and margin parameters for the square and ball are the same as in the first experiment, and the depth for each tunnel is 2 m. As shown in Fig. \ref{fig_exp_tunnel}, the drone successfully completes the race within 5.56 s, with an average tracking RMSE of 0.15 m. The maximum speed reached 6.83 m/s, which is a considerable value in such a small environment. Fig. \ref{fig_exp_tunnel_data} plots the trajectory from a top view. It can be seen that the drone is navigating through the most time-efficient corners within the tunnel, showcasing exceptional spatial utilization in such a narrow race track.

In the final experiment, we demonstrate the applicability of our planner in challenging race tracks. As illustrated in Figure \ref{fig_exp_dive}, the drone is required to dive through the rightmost gate from above and exit through the left face. Accurate modeling of gate constraints is crucial in such scenarios, as collisions can easily happen once the drone dives into the gate. We model the dive gate as a tunnel with two faces. Besides a upper square gate and a left rectangular gate, there are two polyhedron gates to specify the collision-free region inside the dive gate, as shown in Fig. \ref{fig_exp_dive_data}. The results show that our method can consistently accomplish diving flight with a speed of 2.90 m/s during the diving, which validates the remarkable versatility of our method.

\section{CONCLUSIONS}

We present an efficient time-optimal trajectory planner that can faithfully account for the racing gate's shape and size. Compared to existing works, our framework can better utilize the free space of the gates to find the most time-efficient path and thereby yield a comparable time optimality to discretization-based methods. We validate our method via real-world experiments and demonstrate superior performance. In the future work, we will incorporate gates with irregular structures such that all types of gates or corridors appearing in drone racing can be well-modeled.




%

\section*{ACKNOWLEDGMENT}

The authors would like to acknowledge the sponsorship by the Natural Sciences and Engineering Research Council of Canada (NSERC) under grant RGPIN-2023-05148. Additionally, we express our gratitude for the experimental support provided by the Learning Systems \& Robotics Lab (formerly Dynamic Systems Lab) at the UTIAS.



\bibliographystyle{IEEEtran}
\bibliography{IEEEabrv,togt}

\end{document}